\documentclass[conference]{IEEEtran}
\IEEEoverridecommandlockouts
\usepackage{cite}
\usepackage{amsmath,amssymb,amsfonts}
\usepackage{algorithmic}
\usepackage{graphicx}
\usepackage{textcomp}
\usepackage{pifont}
\usepackage{subfig}
\usepackage{url}
\usepackage{multirow}
\newcommand{\cmark}{\ding{51}}%
\newcommand{\xmark}{\ding{55}}%
\usepackage{diagbox}\usepackage{xcolor}
\def\BibTeX{{\rm B\kern-.05em{\sc i\kern-.025em b}\kern-.08em
    T\kern-.1667em\lower.7ex\hbox{E}\kern-.125emX}}
\begin{document}

\title{Deep Ensemble Art Style Recognition
\thanks{This research has been co‐financed by the European Union and Greek national funds through the Operational Program Competitiveness, Entrepreneurship and Innovation, under the call RESEARCH – CREATE – INNOVATE (project code:T1EDK-01728 – Project Acronym: ANTIKLEIA)}
}
\author{\IEEEauthorblockN{Orfeas Menis - Mastromichalakis}
\IEEEauthorblockA{\textit{School of Electrical and Computer Eng.} \\
\textit{National Technical University of Athens}\\
Athens, Greece \\
menorf@islab.ntua.gr}
\and
\IEEEauthorblockN{Natasa Sofou}
\IEEEauthorblockA{\textit{School of Electrical and Computer Eng.} \\
\textit{National Technical University of Athens}\\
Athens, Greece \\
natasa@image.ntua.gr}
\and
\IEEEauthorblockN{Giorgos Stamou}
\IEEEauthorblockA{\textit{School of Electrical and Computer Eng.} \\
\textit{National Technical University of Athens}\\
Athens, Greece \\
gstam@cs.ntua.gr}

}

\maketitle

\begin{abstract}
The massive digitization of artworks during the last decades created the need for categorization, analysis, and management of huge amounts of data related to abstract concepts, highlighting a challenging problem in the field of computer science. The rapid progress of artificial intelligence and neural networks has provided tools and technologies that seem worthy of the challenge. Recognition of various art features in artworks has gained attention in the deep learning society. In this paper, we are concerned with the problem of art style recognition using deep networks. We compare the performance of 8 different deep architectures (VGG16, VGG19, ResNet50, ResNet152, Inception-V3, DenseNet121, DenseNet201 and Inception-ResNet-V2), on two different art datasets, including 3 architectures that have never been used on this task before, leading to state-of-the-art performance. We study the effect of data preprocessing prior to applying a deep learning model. We introduce a stacking ensemble method combining the results of first-stage classifiers through a meta-classifier, with the innovation of a versatile approach based on multiple models that extract and recognize different characteristics of the input, creating a more consistent model compared to existing works and achieving state-of-the-art accuracy on the largest art dataset available (WikiArt - 68,55\%). We also discuss the impact of the data and art styles themselves on the performance of our models forming a manifold perspective on the problem. 

\end{abstract}


\section{Introduction}
In recent years we have witnessed a massive digitization of our world. Art, being an integral part of human culture, has entered its digital era. Museums, galleries, art centers, even private art collectors have digitized their collections to preserve, analyze and sometimes make them publicly available. This constantly growing amount of data imposes the need for automatic classification and analysis of the digitized artworks. 
Artificial intelligence offers powerful tools to solve such problems that require human intuition and intelligence. 

Art style recognition has been traditionally performed by art historians and curators. The last decade there have been attempts to automate the task, achieving notable results. Most successful approaches of the problem employ convolutional neural networks in combination with transfer learning. Transfer learning allows the use of pre-existing knowledge obtained from a similar problem to be employed to deal with a more complex one, usually with less data available. In this paper, we fine-tune networks that were pre-trained on object recognition using the ImageNet Large-Scale Visual Recognition Challenge (ILSVRC) dataset \cite{10.1007/s11263-015-0816-y} of more than 1.2 million natural images of objects with 1,000 categories.

In visual arts, style is a "...distinctive manner which permits the grouping of works into related categories"\cite{Style}. The art style of an artwork is defined only by its appearance. It is determined by the characteristics that describe the artwork, such as the colors and content. Artworks that share certain common features are considered to have the same style. 

\begin{figure}[t]
  \centering
  \subfloat["Lady with an Ermine", Leonardo da Vinci]{\includegraphics[width=0.22\textwidth]{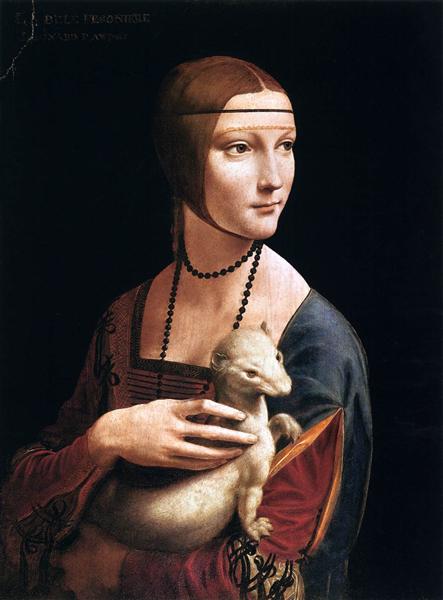}}
  \hfill
  \subfloat["Girl with a Pearl Earring", Johannes Vermeer]{\includegraphics[width=0.255\textwidth]{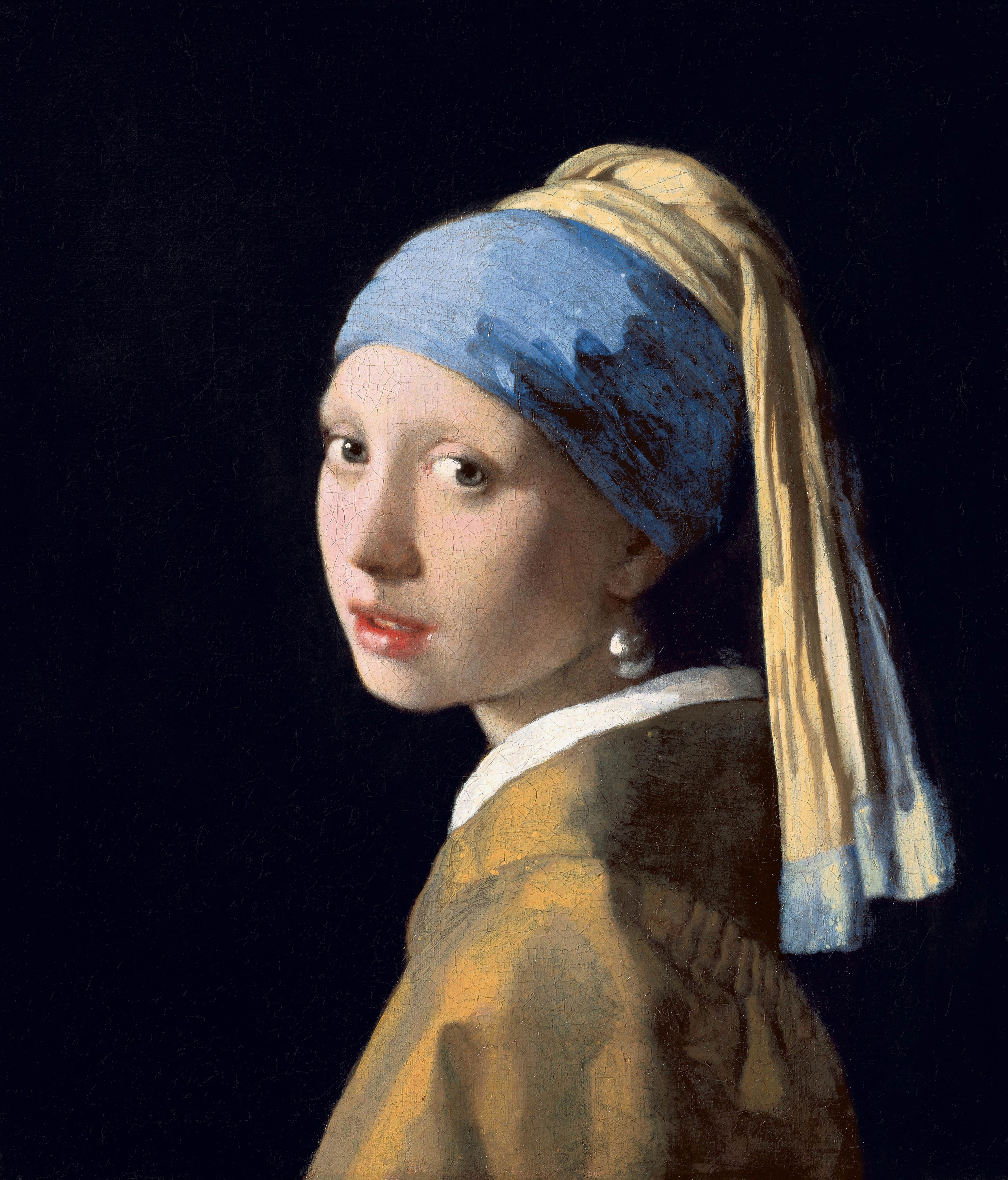}}
  \hfill
 \caption{Similar paintings-Different styles}
 \label{fig4}
\end{figure}

Art style recognition is not a trivial problem like object recognition. 
Art movements inherited features from their predecessors and influenced their successors, creating an amalgam of different characteristics from various styles that formed new ones. Many artworks have elements of multiple art styles, different art styles share common characteristics, artists influenced by different art movements throughout their lives, created pieces of different styles that shared the artist’s personal style. In Fig.~\ref{fig4} Leonardo da Vinci's "Lady with an Ermine" and Johannes Vermeer's "Girl with a Pearl Earring" are illustrated. They both depict a young woman with similar color tones and blank and dark background. There are no sharp edges, while shadows create the illusion of a third dimension. Although the two paintings share many common characteristics they have different art styles. Da Vinci's masterpiece is part of the High Renaissance movement, while the "Girl with a Pearl Earring" is a typical Baroque painting. Therefore, defining an artwork's unique art style can be a very challenging task, even for an expert.

Art is an important source for history, sociology and various other scientific areas. The automatic analysis and classification of the huge amount of artworks that is digitally available, can accelerate significantly the knowledge extraction procedure, leading to areas that have never been explored before. Interestingly, there have not been many adequate attempts to exploit all that knowledge through machines. It is a relatively new area of study for machine learning, since only during the last decade a sufficient amount of data was made available. Although Deep Learning has surpassed human level performance on object recognition \cite{he2015delving}, it has reached a crucial point on art recognition where traditional methods fail to improve the performance of the models. Therefore, researchers have to develop methodologies that meet the particular needs of art recognition and in parallel exploit the tools and technologies that have been developed for image recognition. The main purpose of this study was not only to adjust existing architectures and techniques to the specific problem of art recognition, but also to use that knowledge to introduce new methodologies that adapt to its special needs. More specifically, various models were tested achieving state-of-the-art performance and employing 3 architectures (DenseNet121, DenseNet201 and Inception-ResNet-V2), that to the best of our knowledge have never been used before on art recognition. A stacking ensemble method is introduced, creating a super-model composed of models that recognize different characteristics of the input and complement one another by eliminating each other’s weaknesses. Other ensemble solutions fed patches of images to identical networks in order to extract more information from the data \cite{8631731,8675906}, resulting to data-dependent issues and significant differences on the performance between datasets. We overcame that problem by employing sub-models with slightly different perception of art style and created more robust and consistent models. The proposed model took the lead on the most challenging dataset, the WikiArt collection, and achieved the second best accuracy on Pandora 18K collection.
 
\section{Related Work}
As AI continuously progresses, problems from areas with more vague concepts that require human intelligence can be addressed effectively. Art is the apogee of human creation as well as an integral part of human culture. Areas of art like painting and music \cite{oord2016wavenet,huang2016deep} have been the target of many researchers, who have been trying to classify and analyze data, authenticate forgeries \cite{5413338} or even create artworks \cite{gatys2015neural,johnson2016perceptual}. In this work we focus on the problem of art style recognition and classification. 
Early studies addressed the problem with traditional machine learning methods based on features extraction from images of datasets containing a relatively small number of artworks.
Jia Li and Wang \cite{1278358} used two-dimensional multiresolution hidden Markov models (MHMMs) to classify the artworks of 5 Chinese artists. Gunsel et al. \cite{Gunsel} approached the problem of painter and art movement recognition with Bayesian, SVM and kNN classifiers. SVMs were also employed by Jiang et al. \cite{Jiang} who used low-level features like color histograms, color coherence vectors and autocorrelation texture features to detect traditional chinese paintings from general images and categorize them into Gongbi and Xieyi schools. More studies approached the problem of art recognition from different aspects and with different tools\cite{5403040,Shamir:2010:IES:1670671.1670672,Khan2014,Arora2012TowardsAC}, but they all lacked a proper dataset. The datasets that were used had either very few artworks or very few classes. Karayev et al. \cite{karayev2013recognizing} carried out the first large-scale study on art recognition and created an impressively large, publicly available dataset with artworks they collected from the visual art encyclopedia WikiArt \footnote{"WikiArt, Visual Art Encyclopedia", www.wikiart.org}. They approached the problem through the field of deep learning using convolutional neural networks, which since then, has been the best and most widely used approach to the problem. Bar et al. \cite{Bar2014ClassificationOA} and Peng and Chen \cite{7351365} also used convolutional filters pre-trained on ImageNet as proposed by Donahue et al. \cite{donahue2013decaf} . In 2017, Florea et al. \cite{7926652} created the Paintings Dataset for Recognizing the Art Movement (Pandora 18K) collection, which is a relatively small dataset verified by experts. They approached the task of art recognition with extracted features and SVMs as well as with the use of CNNs and pre-trained networks with architectures like AlexNet \cite{AlexNet}. Since then, Pandora 18K is widely used in art recognition studies. Lecoutre et al. \cite{Lecoutre2017RecognizingAS} achieved an improvement in accuracy with a deeper training of a pre-trained ResNet based model, which showed that although object recognition models can serve as a starting point for art recognition, the tasks are not that similar so as to use frozen parts of a network recognizing objects, to recognize art. In 2018, Rodriguez et al. \cite{8631731} introduced a method to divide the artwork into five patches, classify each patch independently and then combine the outputs with a weighted sum to get the final output. Sandoval et al. \cite{8675906} went one step further and trained a shallow neural network on top of the previously introduced patch-based classifier. This increased impressively the accuracy of their model which achieved the state-of-the-art results. Our proposed methodology presented in this paper, attempts to successfully surpass their accuracy on WikiArt dataset and approach their Pandora 18K accuracy.

\section{Datasets}
\label{Datasets}
Two datasets of digital images of artworks collated from publicly available fine art collections were used to train and evaluate our models. We assessed both datasets to be adequate for this task. 

\subsection{Dataset 1}
\begin{figure}[b]
\centerline{\includegraphics[width=0.5\textwidth]{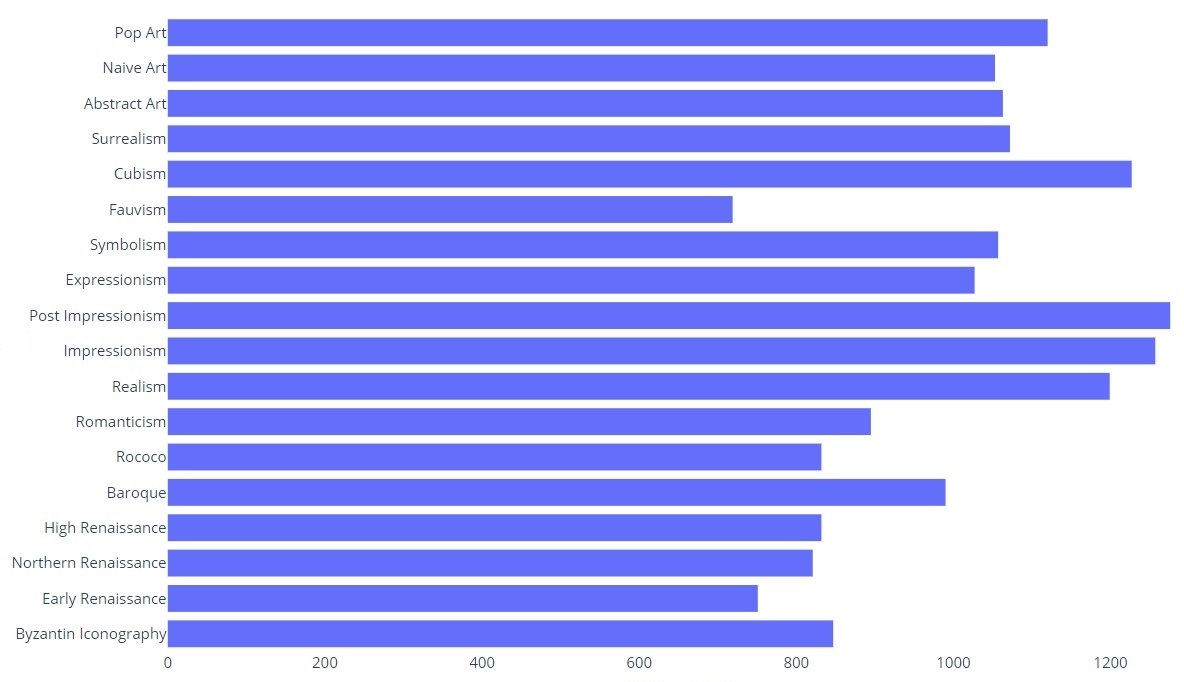}}
\caption{Number of artworks per class in Dataset 1.}
\label{fig5}
\end{figure}

Dataset 1 is the Paintings Dataset for Recognizing the Art Movement (Pandora 18K) collection. It consists of 18038 paintings divided into 18 categories (Byzantine Iconography, Early Renaissance, Northern Renaissance, High Renaissance, Baroque, Rococo, Romanticism, Realism, Impressionism, Post Impressionism, Expressionism, Symbolism, Fauvism, Cubism, Surrealism, Abstract Art, Naive Art and Pop Art). Each label’s validity has been verified by engineers and art experts. In Fig.~\ref{fig5} we present the distribution of artworks in the classes.

\subsection{Dataset 2}
\begin{figure}[b]
\centerline{\includegraphics[width=0.5\textwidth]{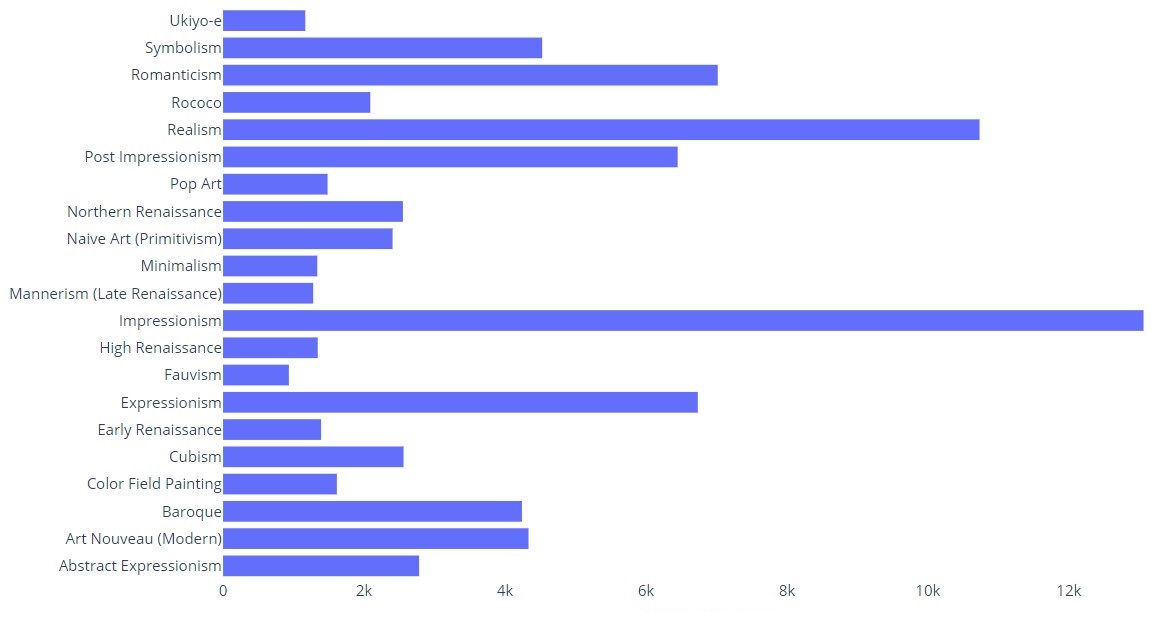}}
\caption{Number of artworks per class in Dataset 2.}
\label{fig6}
\end{figure}

Dataset 2 consists of artworks from the visual art encyclopedia “wikiart.org”. The WikiArt collection was first collected by Karayev et al. \cite{karayev2013recognizing} in 2013 and it was the first large-scale dataset used for automatic art classification. The original WikiArt dataset used, came from the repository in \cite{ArtGAN}. It originally contained 81445 digital images of artworks, divided into 27 categories (Abstract Expressionism, Action Painting, Analytical Cubism, Art Nouveau (Modern), Baroque, Color Field Painting, Contemporary Realism, Cubism, Early Renaissance, Expressionism, Fauvism, High Renaissance, Impressionism, Mannerism (Late Renaissance), Minimalism, Naive Art (Primitivism), New Realism, Northern Renaissance, Pointillism, Pop Art, Post Impressionism, Realism, Rococo, Romanticism, Symbolism, Synthetic Cubism and Ukiyo-e). The WikiArt collection has two main disadvantages: (a) it is an imbalanced set and (b) its data are labeled by users and not by art experts like in the Pandora 18K. We examined the set prior to use and we selected 23 out of the 27 categories and merged three categories, which were part of the same art movement and style. The selection was made in order to achieve a more balanced and authoritative dataset. Analytical Cubism, Synthetic Cubism and Cubism were merged into the Cubism class since they are all parts of the same art movement and style. Action Painting is considered by some art experts to be part of Abstract Expressionism, but since it is unclear with very few data we decided to exclude it from our study. Under same rationale, we excluded New Realism, Contemporary Realism and Pointillism either because they had too few data or because they were not broadly accepted as a separate art style. The resulting WikiArt dataset has 80039 digital images of artworks divided into 21 categories. Fig.~\ref{fig6} illustrates the number of images per class. 

The use of two different datasets aimed to demonstrate and prove that the proposed models generalize well and are dataset-independent.

\section{Method}
\label{method}
In the current study we train different CNNs with the method of transfer learning and we fine-tune the hyper-parameters on single-architecture models. Then we apply a form of ensemble learning where we combine the knowledge of our models in a meta-classifier leading to improved results. We use Keras \cite{chollet2015keras} on top of Tensorflow \cite{tensorflow2015-whitepaper} to develop our models. The weights of the pre-trained networks are provided by Keras. 

Data Augmentation is applied to prevent overfitting. At every epoch, the data is fed to the network slightly altered with some distortions applied with a predefined probability. This aims to alter the input so as to prevent the model from learning specific spatial relations of the images, which would lead to overfitting, but at the same time retain the characteristics that define the class of the input.
For validation and evaluation purposes we split the datasets into train, validation and test sets, keeping the distribution of classes the same in every set. 80\% of the data were used for training, 10\% for validation and 10\% for testing. We used accuracy as the evaluation metric. 
\subsection{Baseline}
\label{MethodBaseline}
 As a first stage, we train simple models to be used as base for the next steps. We loaded weights of networks pre-trained to recognize objects on the ImageNet Large-Scale Visual Recognition Challenge (ILSVRC) dataset. Then, the full network is trained on the art dataset changing the top layers to suit the problem requirements. The number of nodes of the last dense layer depends on the number of the classes of each dataset. We tried to keep the architectures intact and make the minimum changes possible so that we take the most out of them but also adapt them to our requirements. Throughout the training process, the behavior of the models with different hyper-parameters values were examined in combination with different techniques of data preprocessing. 

\subsection{Simple Ensemble}
\label{MethodSimple}
After setting the baseline, the outputs of different individual already trained networks were combined, to create a super-model. We tried to detect and use good-performing models that perceived a slightly different perspective of art style so that one complements the other. We used the simple models of section \ref{MethodBaseline} to compose a more powerful classifier. We created a custom data generator to feed multiple versions of the input image to the individual networks adapted to their requirements. For each image we combined the output of the sub-classifiers by calculating either the average, the maximum or the minimum of each class value that formed the output of the super-model. This method does not require extra training.

\subsection{Stacking Ensemble}
\begin{figure}[t]
\centerline{\includegraphics[width=0.5\textwidth]{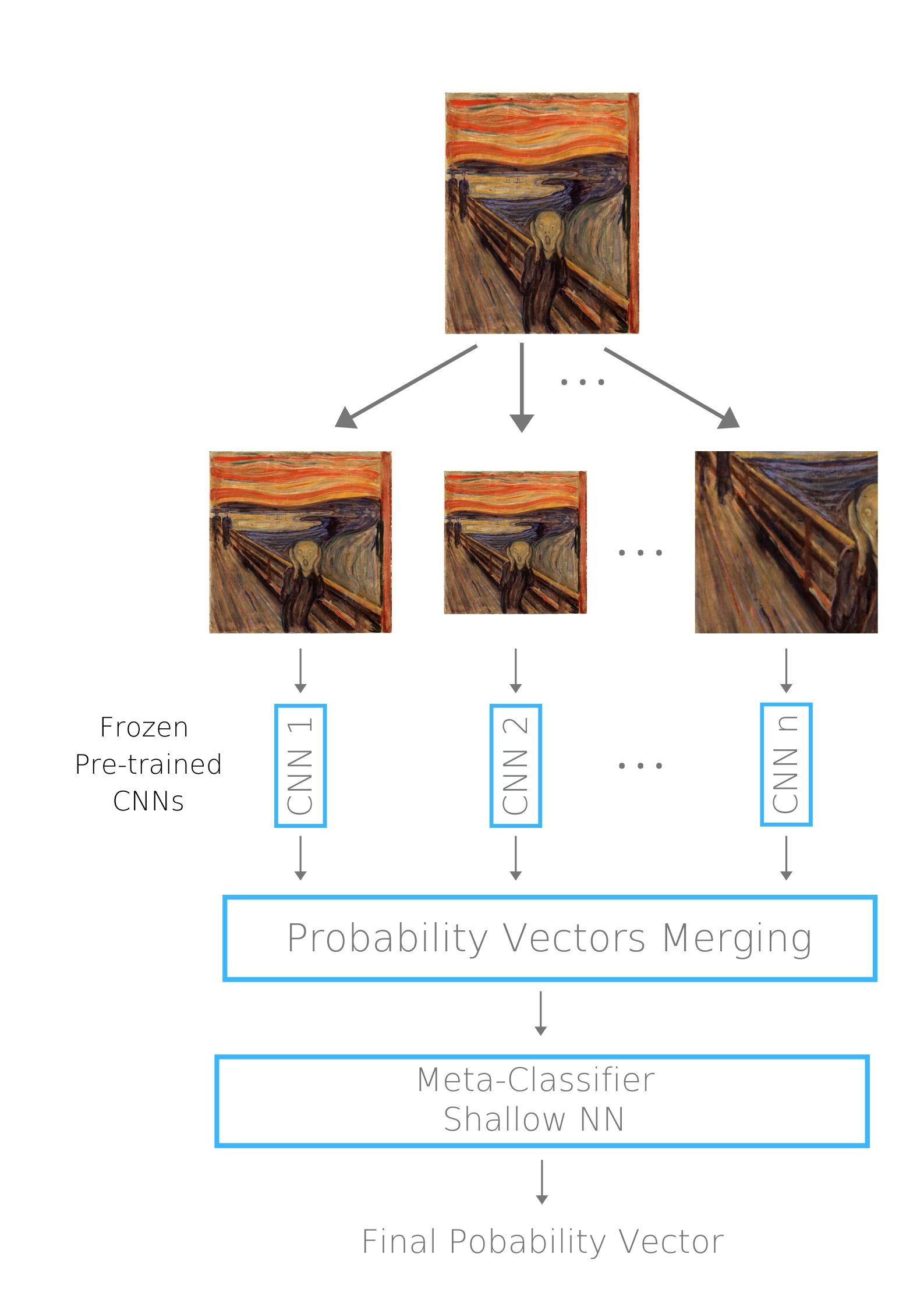}}
\caption{The proposed stacking ensemble method.}
\label{fig1}
\end{figure}

To combine more effectively the knowledge of the individual models, a shallow neural network was trained on top of the sub-models. The first stage of our proposed stacking ensemble method is the same with the simple ensemble described in section \ref{MethodSimple}. The same custom data generator is used to feed the individual networks with the proper form of each input image. Then the outputs of the models are merged into a vector of size $N\times M$, where N is the number of sub-models and M the number of classes of the dataset. That vector is then fed to a shallow neural network used as a meta-classifier, trained to properly combine the merged outputs and determine the art style of the initial input image, as shown in Fig.~\ref{fig1}. The individual base models are frozen so as not to get affected by the meta-classifier's training process. Different kinds of shallow neural networks were tested, with one or two dense layers and intermediate dropout layers.

\section{Experimental Results}
The experiments were conducted on a GeForce GTX 1080 Ti GPU.

\subsection{Setting the baseline}
\label{SettingBaseline}
The first part of this work illustrates a comparative study between different architectures, methods of data preprocessing (resizing and cropping) and levels of data augmentation, highlighting the way that each option impacts the training process and the results.

\subsubsection{Data Preprocessing}
As data are vital for every machine learning model, a deeper understanding of the data can significantly improve the performance of a model. Every convolutional neural network has specific requirements for the size of the input images. The datasets used in this work consist of digital artwork images of different sizes and shapes. Therefore, images were configured to fit each architecture’s requirements by cropping and/or resizing. Most of the experiments were conducted by resizing each image to the required size, but we also experimented with feeding the center or a random crop of each image to the network. When cropping an image, the pixel density is kept intact preserving the brushstrokes and pixel-level details of the painting. On the other hand, by resizing some of the artwork details might be destroyed. However, since the whole artwork is fed to the network, higher-level features like human or animal shapes are preserved, and thus the model is assisted in identifying the correct class. This is mainly due to the fact that the content of an artwork is sometimes related to its style. For example, it is much more likely to find a person in a baroque painting rather than an abstract expressionism one. 

\begin{table}[t]
\caption{Resize-Crop}
\begin{center}
\begin{tabular}{|c|c|c|}
\hline
\backslashbox{Model}{Data}&\textbf{Resized Images}&\textbf{Cropped Images}  \\
\hline
\textbf{Trained on Resized} & {66,56\%} & {57,02\%} \\
\hline
\textbf{Trained on Cropped} & {62,84\%} & {65,06\%} \\
\hline
\end{tabular}
\label{tab1}
\end{center}
\end{table}

Models that were trained on resized images performed slightly better (1,5\%) than models who were trained on cropped images. A very interesting observation is how the accuracy of each model is affected when tested on data that have undergone different preprocessing, i.e. a model trained on resized images gets tested on cropped images and vice versa. The model trained on cropped images performs equitably or somewhat worse on resized images but on the other hand, the model trained on resized images has almost 10\% less accuracy on cropped images. This leads us to two elementary conclusions. Firstly, the content of an artwork is probably considered by the deep learning model as a characteristic of the art piece’s style, which might be responsible for the better accuracy achieved on the resized images. That element is absent in the cropped images, hence the worse performance. Secondly, the art style is present also in parts of an artwork allowing the classifier to understand a more general aspect of each style when trained on patches of the image, qualifying it to recognize art style in cropped as well as in resized images. It seems that each model learns a slightly different aspect of art style, which could be very useful when combining the predictions of different models. 

\begin{figure}[b]
  \centering
  \subfloat[Level 1]{\includegraphics[width=0.24\textwidth]{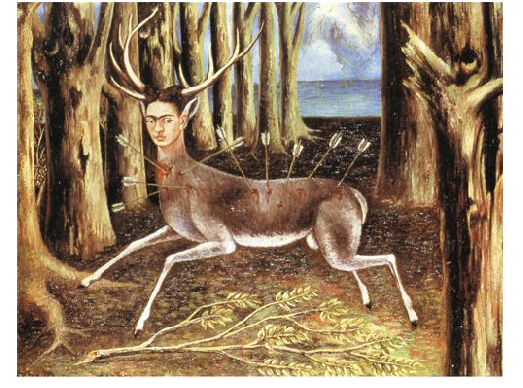}}
  \hfill
  \subfloat[Level 2]{\includegraphics[width=0.24\textwidth]{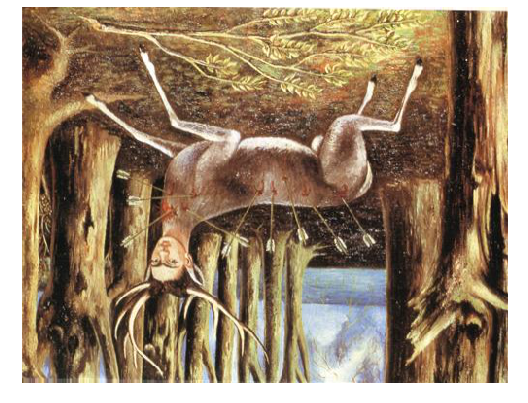}}
  \hfill
  \subfloat[Level 3]{\includegraphics[width=0.24\textwidth]{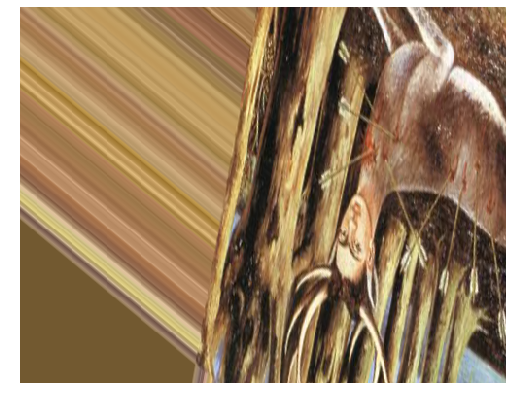}}
  \hfill
  \subfloat[Level 4]{\includegraphics[width=0.24\textwidth]{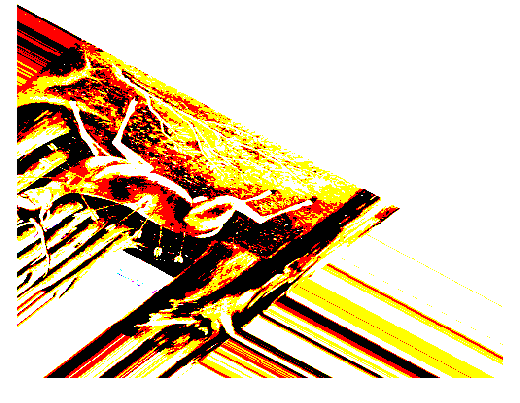}}
  \hfill
 \caption{Data augmentation levels on a painting}
 \label{fig2}
\end{figure}

Overfitting prevents a deep learning model from achieving higher levels of accuracy, since the network’s weights overly adapt to the training data leading to worse generalization to unseen data. We addressed this problem through data augmentation. The form of data augmentation used, slightly alters the input images at every epoch so that the model does not learn specific space related features, but learns general characteristics of each class. This improved the performance of the proposed models achieving higher accuracy and less overfitting. Experiments with different levels of data augmentation were conducted, with each level applying more and/or intenser distortions. Data augmentation should be used cautiously since too much distortion could corrupt the defining class characteristics of each image, making the model incapable of training. In table \ref{tab2} the distortions applied at each level of data augmentation are presented and in Fig.~\ref{fig2} the results of these distortions on Frida Khalo’s painting “The wounded deer” are illustrated. 

\begin{table}[b]
\caption{Data Augmentation Levels}
\begin{center}
\begin{tabular}{|c|c|c|c|c|}
\hline
\backslashbox{Distortion}{Level}&\textbf{Level 1}&\textbf{Level 2}& \textbf{Level 3}&\textbf{Level 4}  \\
\hline
\textbf{Channel shift} & {\xmark} & {\xmark} & {\xmark}  & {\cmark}\\
\hline
\textbf{Brightness change} & {\xmark} & {\xmark} & {\xmark} & {\cmark}\\
\hline
\textbf{rotation} &{\xmark} & {\xmark} & {\cmark}& {\cmark}\\
\hline
\textbf{width shift} &{\xmark} & {\xmark} & {\cmark}& {\cmark}\\
\hline
\textbf{height shift} &{\xmark} & {\xmark} & {\cmark}& {\cmark}\\
\hline
\textbf{zoom} &{\xmark} & {\xmark} & {\cmark}& {\cmark}\\
\hline
\textbf{shear} &{\xmark} & {\cmark} & {\cmark}& {\cmark}\\
\hline
\textbf{horizontal flip} &{\xmark} & {\cmark} & {\cmark}& {\cmark}\\
\hline
\textbf{vertical flip} &{\xmark} & {\cmark} & {\cmark}& {\cmark}\\
\hline
\end{tabular}
\label{tab2}
\end{center}
\end{table}

Table \ref{tab3} illustrates the accuracy of a DenseNet121 model trained on Dataset 1, when applying the four different levels of data augmentation mentioned above, during the training process.

\begin{table}[b]
\caption{Data Augmentation Levels - Accuracy}
\begin{center}
\begin{tabular}{|c|c|c|c|}
\hline
\backslashbox{Level}{Accuracy}&\textbf{Train Set}&\textbf{Validation Set}&\textbf{Test Set}\\
\hline
\textbf{Level 1} & {99,70\%} & {62,72\%} & {61,06\%}\\
\hline
\textbf{Level 2} & {91,13\%} & {65,02\%} & {63,12\%}\\
\hline
\textbf{Level 3} & {74,00\%} & {65,13\%} & {64,23\%} \\
\hline
\textbf{Level 4} & {37,01\%} & {50,83\%} & {50,49\%}\\
\hline
\end{tabular}
\label{tab3}
\end{center}
\end{table}
As expected, training accuracy drops as data augmentation increases. As it was noticed, testing accuracy, which is what we are trying to improve, increases along with data augmentation up to level 3 and then drops dramatically at level 4. Accuracy drops due to the fact that at level 4, as seen in Fig.~\ref{fig2}, the distortions deform the artwork and make it almost unrecognizable, thus constraining the network to learn the real representation of each class. On the other hand, up to level 3 the alterations prevent the model from associating specific data and details with styles, thus reducing train accuracy, and enabling the system to learn a more general representation of each style, increasing the accuracy on unseen data (test accuracy). As a conclusion, some methods that are counter-intuitive for humans like distorting an image, can assist the deep learning model in learning better the characteristics of the image and improving its performance. However, the impact of such choices should be considered in order to eliminate undesired results. 

\subsubsection{Architectures}
After determining the best suited data preprocessing methods, the performances of 8 different architectures using these methods are compared, in order to create more powerful models that perceive the different aspects of art style and will help us create a state-of-the-art super-model when combined. 

The architecture of a network is one of the most definitive aspects for the model’s success. The architecture defines the form, size and depth of the network, which determines the training procedure and also addresses deep learning problems, like the vanishing gradient problem, overfitting etc.
We trained networks with various architectures that were designed for image recognition and present proven performance near or above the state of the art.
Although state-of-the-art architectures on object recognition tasks don’t perform necessarily the same way on art recognition problems, they serve as a good point to start. Therefore, we chose to train and evaluate networks with the following architectures that present state-of-the-art performance and demonstrate powerful but different characteristics: VGG16, VGG19 \cite{simonyan2014deep}, ResNet50, ResNet152 \cite{he2015deep} , Inception-V3 \cite{szegedy2015rethinking} , Inception-ResNet-V2 \cite{szegedy2016inceptionv4} , DenseNet121 and DenseNet201 \cite{huang2016densely} . The experiments were performed on Dataset 1 and accuracy was used as a metric for evaluation. 

\begin{table}[b]
\caption{Architectures and Dataset 1 Performance}
\begin{center}
\scalebox{0.82}{
\begin{tabular}{|c|c|c|c|c|}
\hline
\textbf{Architecture}&\textbf{Parameters}&\textbf{Layers}&\textbf{Input}&\textbf{Test Accuracy}\\
\hline
\hline
\textbf{VGG 16}&{14.723.922}&{16}&{$224\times224$}&{56,13\%}\\
\hline
\textbf{VGG 19}&{20.033.618}&{19}&{$224\times224$}&{55,97\%}\\
\hline
\textbf{ResNet50}&{23.624.594}&{50}&{$224\times224$}&{59,84\%}\\
\hline
\textbf{ResNet152}&{58.407.826}&{152}&{$224\times224$}&{59,07\%}\\
\hline
\textbf{DenseNet 121}&{7.055.954}&{121}&{$224\times224$}&{64,23\%}\\
\hline
\textbf{DenseNet 201}&{18,321,984}&{201}&{$224\times224$}&{64,67\%}\\
\hline
\textbf{Inception-V3}&{21.839.666}&{48}&{$299\times299$}&{67,55\%}\\
\hline
\textbf{Inception-ResNet-V2}&{54.364.402}&{164}&{$299\times299$}&{66,56\%}\\
\hline
\end{tabular}}
\label{tab4}
\end{center}
\end{table}

The VGG architectures (VGG16 and VGG19) were the oldest and shallowest, and expectedly performed worse than most of the rest architectures. The ResNet architectures (ResNet50 and ResNet152) did not meet the expectations as they did not manage to approach state-of-the-art accuracy, both achieving accuracy below 60\%. The most interesting part of this section of the study is that three (DenseNet121, DenseNet201 and Inception-ResNet-V2) out of the four architectures that managed to surpass the accuracy limit of 60\% and gave near to the state of the art results, to the best of our knowledge, had never been used before on a similar task. Inception-V3 achieved the top results with 67,55\% accuracy, confirming that it is one of the best architectures for art recognition. DenseNet architectures (DenseNet121 and DenseNet201) achieved notable performance with a significant smaller amount of parameters for the shallower version, which leads to a considerable decrease in training time and memory demands. DenseNet121 has the potential to become a powerful tool for systems with memory or processing power limitations. Last but not least, Inception-ResNet-V2 is one of the deepest and largest architectures tested in this study which achieved the second best accuracy (66,56\%) on Dataset 1 and the top accuracy (65,01\%) on Dataset 2. In both datasets Inception-V3 and Inception-ResNet-V2 produced the top results.

\subsection{Proposed Methodology}
The appropriate preprocessing of data along with the extensive study of different architectures and hyper-parameters, derived multiple well-performing models. Art style is a multilevel concept. We had to study and comprehend deeply the underlying concepts of art styles and how they bind with our data. Only then were we able to approach the multiple angles of the problem creating models with different perspectives of art style that can complement one another. Putting together models that achieve high accuracy and combining their outputs does not guarantee better results. Models with different characteristics and different approaches to the problem must be chosen, so that each one can contribute to the final result. This is the focus of the second part of our study, where we combine the appropriate models derived in the first part. We set as baseline the performance of the individual CNNs that were trained in section \ref{SettingBaseline}. We examined closely our models’ characteristics and outputs. Confusion Matrices were used to provide a view of the classes on which each model illustrates not only better performance but also weaknesses. The methods described in the section \ref{method} were implemented to bring together the parts of knowledge that we had already collected. We managed to reach accuracy just above 70\% on Dataset 1 with the first simple ensemble method and state-of-the-art results on Dataset 2 with 67,92\% accuracy on the test set. The output of this method is the result of a per-class calculation of either the minimum, the maximum or the average. Experimental results show that there is no better choice of the three, they all lead to similar results with the top choice depending on the sub-models. In table \ref{tab5} we present the results of the proposed methodology on Datasets 1 and 2. 

\begin{table}[b]
\caption{Results}
\begin{center}
\begin{tabular}{|c|c|c|}
\hline
\backslashbox{Model}{Test Set Accuracy}&\textbf{Dataset 1}&\textbf{Dataset 2}\\
\hline
\textbf{Baseline - Single CNN} & {67,55\%} & {65,01\%} \\
\hline
\textbf{Simple Ensemble} & {70,33\%} & {67,92\%} \\
\hline
\textbf{Stacking Ensemble} & {72,47\%} & {68,55\%} \\
\hline
\end{tabular}
\label{tab5}
\end{center}
\end{table}
As it can be observed, training a shallow neural network to better combine the outputs of our CNNs, improves the results about 2\% on Dataset 1 and 0,6\% on Dataset 2. It is interesting that when models trained on cropped images were added to the stacking ensemble model, it improved the results, although those models had a lower test accuracy compared to models trained on resized images. This supports our claim that we do not only need powerful classifiers but also models that conceive different aspects of art style. The models used to achieve top results on Dataset 1 consist of Inception-V3 and Inception-ResNet-V2 classifiers, a model trained with class weights to improve the performance on weak classes, and two models trained on cropped images, one with the center part of each image as input and the second with a random part cropped, both cropped parts obtained after resizing the input image to the double of the required dimensions (e.g. if the CNN's input size is $299\times299$, first the image is resized to $598\times598$ and then cropped a $299\times299$ patch of it). Same for Dataset 2 but without the model trained with class weights. 

\section{Discussion}
\subsection{Comparative Study}
To evaluate the results we have to compare with earlier studies. In this work with the term accuracy we refer to the test set accuracy. In table \ref{tab6} we present the top accuracy achieved by each study on art style recognition on the Pandora 18K dataset (Dataset 1). Rodriguez et al. \cite{8631731} and Sandoval et al.\cite{8675906} added an extra class with Australian Aboriginal Art. According to their results and the accuracy of that specific category, the accuracy on the Pandora 18K was calculate approximately without the extra class. We also provide the accuracy on the new dataset (Pandora18K + Australian Aboriginal Art) in the parenthesis. 
\begin{table}[b]
\caption{Pandora 18K Accuracy}
\begin{center}
\begin{tabular}{|c|c|}
\hline
{Study}&{Accuracy}\\
\hline
\hline
{Florea et al. \cite{7926652}} & {50,1\%} \\
\hline
{Florea and Gieseke \cite{FLOREA2018220}} & {63,50\%}  \\
\hline
{Rodriguez et al. \cite{8631731}} & {68,10\% (*70,20\%)} \\
\hline
{Sandoval et al. \cite{8675906}} & {76,01\%(*77,53\%)}\\
\hline
\textbf{This work} & \textbf{72,47\%}\\
\hline
\multicolumn{2}{c}{$^{\mathrm{*}}$Accuracy with the Australian Aboriginal Art class.}
\end{tabular}
\label{tab6}
\end{center}
\end{table}

Since there are different versions of the WikiArt dataset, table \ref{tab7} illustrates the accuracy of various studies on art style recognition on the WikiArt dataset (Dataset 2) along with the number of artworks and the number of classes in each case. 

\begin{table}[b]
\caption{WikiArt Accuracy}
\begin{center}
\begin{tabular}{ |c|c|c|c| } 
 \hline
 Study & Accuracy & Classes & Artworks\\
 \hline
 \hline
 Karayev et al. \cite{karayev2013recognizing} & 44,10\% & 25 & 85000\\
 \hline
 Bar et al. \cite{Bar2014ClassificationOA} & 43,02\% & 27 & 40724\\
 \hline
 Saleh and Elgammal \cite{saleh2015largescale} & 46,00\% & 27 & 81449\\
 \hline
 Tan et al. \cite{7533051} & 54,50\%  & 27 & 80000 \\
 \hline
 Lecoutre et al. \cite{Lecoutre2017RecognizingAS} & 62,80\% & 25 & 80000 \\
 \hline
 Florea and Gieseke \cite{FLOREA2018220} & 46,20\% & 25 & 85000 \\
 \hline
 Cetinic et al. \cite{CETINIC2018107} & 56,40\% & 27 & 96014 \\
 \hline
 Zhong et al. \cite{Zhong2019} & 59,01\% & 25 & 30825 \\
 \hline
 Sandoval et al. \cite{8675906} & 66,71\% & 22 & 26400 \\
 \hline
 \textbf{This work} & \textbf{68,55\%} & \textbf{21} & \textbf{80039} \\
 \hline
\end{tabular}
\label{tab7}
\end{center}
\end{table}

The obtained results achieve state-of-the-art performance and take the lead on the most challenging dataset, the WikiArt collection. We also trained models on the initial WikiArt dataset with around 81500 images divided into 27 classes and the results were very similar, so the changes made on WikiArt did not improve the results. Although Sandoval et al. \cite{8675906} achieved an impressive accuracy on Pandora 18K dataset, their models are not consistent, giving more than 10\% less accuracy on their improved WikiArt dataset. We managed to overcome issues of data dependencies by using sub-models with slightly different approaches of art style and created robust and data independent  models.

\subsection{Results Observations}
\begin{figure}[t]
\centerline{\includegraphics[width=0.5\textwidth]{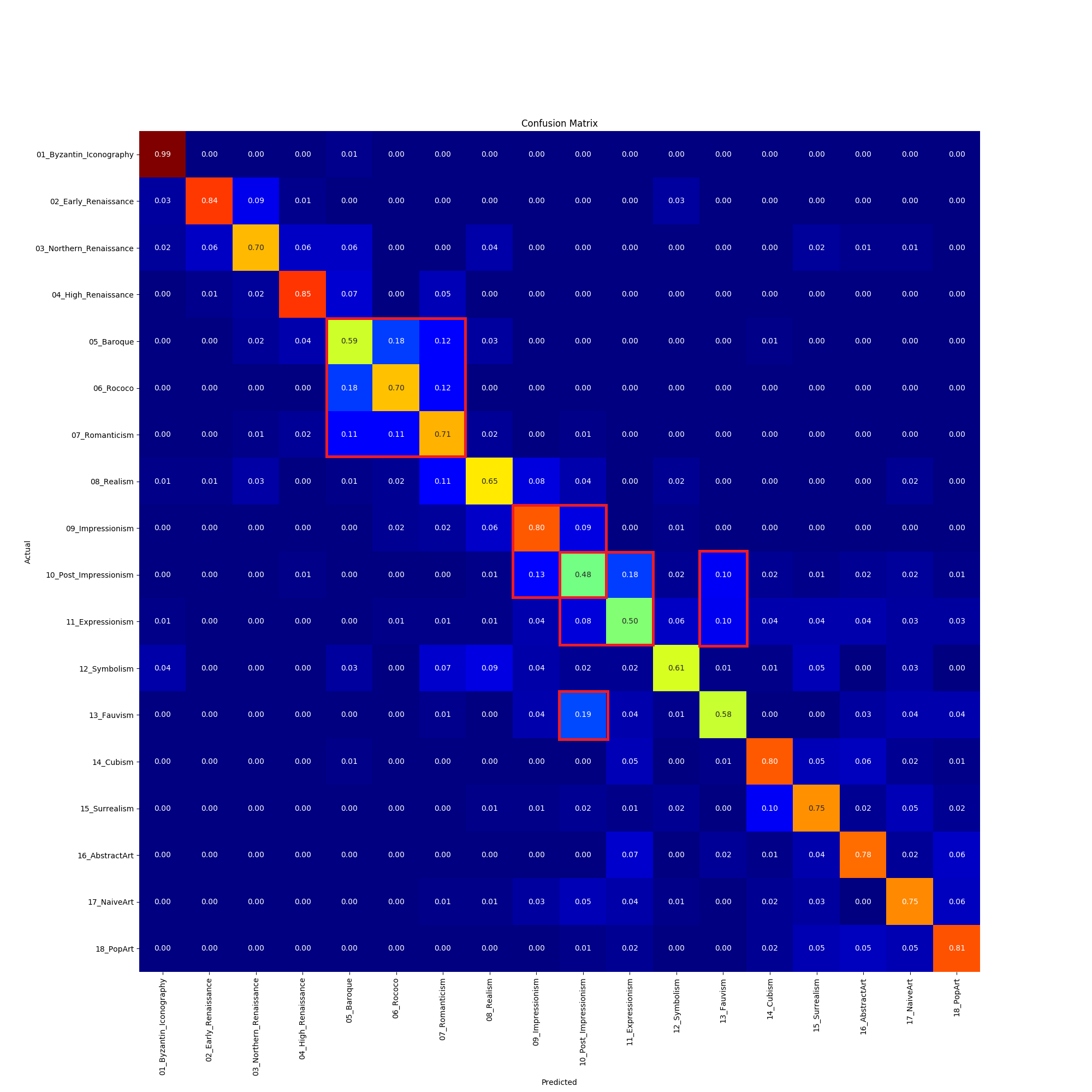}}
\caption{Confusion Matrix - Similar Styles.}
\label{fig3}
\end{figure}

After the completion of experiments, the obtained results were evaluated and explicated so as to establish a multi-faceted perception of the problem and provide an interpretation of the outcome of our study. When examining the Confusion Matrices produced by our models we detected some classes that are more often confused with some of them highlighted in Fig~\ref{fig3} (red boxes). The styles that are confused the most are styles that share common characteristics and often mislead and confuse art experts too. As noticed, Baroque, Rococo and Romanticism are frequently confused, but there is a reason for that. Rococo is sometimes also referred as “Late Baroque” since it arose at the end of the Baroque period in the same areas of Europe, and adopted many of its characteristics. The movement of Romanticism was highly influenced by artists of the Late Baroque period and inherited some of the Baroque elements from their Rococo artworks. Accordingly, Impressionism and Post-Impressionism share many common characteristics, as well as Post-Impressionism and Expressionism. These are art movements that one succeeded the other through the works of artists that were influenced by different movements of their period, forming new styles on the base of older ones. These relations are not always easy to derive, but in some cases elements of more than one style are present making it very difficult to label an artwork with only one style.
The separating lines between art styles are blurry creating debates even between experts. An AI system perceives and recognizes an artwork in a similar manner that humans do. It can detect connections between different styles, which could be considered as a very interesting aspect of our problem. It can also extract useful information even from the misclassifications, sometimes leading to even more fascinating discoveries. Connections of the creators of artworks to unknown artist can be derived, establishing connections that have not been previously assumed or attempted.

On the other end, there are classes that consistently achieve top accuracy even at models that don’t perform very well. Byzantine Iconography in Dataset 1 and Ukiyo-e in Dataset 2 are the most extreme examples demonstrated in our experiments. The accuracy at these two classes is always above 90\% with Byzantine Iconography reaching 99\% in most cases. This is due to the fact that both are art styles very different than the rest in dataset. However, in Byzantine Iconography there might be an extra element which increases the recognition rate. Since it is by far the oldest style in Dataset 1 and many of the artworks are paintings on walls or wood, corrosion is very intense in most pieces, which is perceived by the system as a characteristic of the class. We always have to be aware of the peculiarities of our data because sometimes it can affect our experiments. We don’t consider this particular characteristic important enough to corrupt our results but there are cases where such details may affect the quality and the validity of the data.

\section{Conclusion}
Through this study we presented the importance of understanding the data and all the aspects of the problem. We created a powerful and robust classifier by combining multiple models, with different perspectives of art style. Particular cases where the proposed models were confused between art styles, were found to be historically and artistically consistent, leading to a new perspective on deep learning classifiers' role and capabilities. The proposed models do not aim to replace art experts, but to consult and collaborate with them. We intend to inspect further the different shallow neural networks that can be used as meta-classifiers for our models. We aspire to further study and deploy the proposed models in a more comprehensive way than one-label classification, which will be applicable also to different areas of study with vague concepts and abstruse ideas, like art.

\bibliography{bibliography}
\bibliographystyle{ieeetr}

\end{document}